\documentclass[10pt,twocolumn,letterpaper]{article}

\usepackage[pagenumbers]{cvpr} %
\usepackage[accsupp]{axessibility}

\definecolor{cvprblue}{rgb}{0.21,0.49,0.74}
\usepackage[pagebackref,breaklinks,colorlinks,allcolors=cvprblue]{hyperref}
\usepackage{amsmath,amssymb}
\usepackage{algorithm}
\usepackage{algpseudocode}
\usepackage{makecell}

\def\paperID{***} %
\def\confName{CVPR}
\def\confYear{2026}

\title{On Exact Editing of Flow-Based Diffusion Models}

\author{
  Zixiang Li\textsuperscript{1,2,3}, 
  Yue Song\textsuperscript{4}, 
  Jianing Peng\textsuperscript{1,2,3}, 
  Ting Liu\textsuperscript{3}, 
  Jun Huang\textsuperscript{3},
  Xiaochao Qu\textsuperscript{3},
  Luoqi Liu\textsuperscript{3}, \\
  Wei Wang\textsuperscript{1,2},
  Yao Zhao\textsuperscript{1,2},
  Yunchao Wei\textsuperscript{1,2} \\ 
 \textsuperscript{1}Institute of Information Science, Beijing Jiaotong University \\
  \textsuperscript{2}Visual Intelligence +X International Cooperation Joint Laboratory of MOE \\
  \textsuperscript{3}MT Lab, Meitu Inc 
  \textsuperscript{4}Tsinghua University \\
  }

\begin{document}
\twocolumn[{%
\renewcommand\twocolumn[1][]{#1}%
\maketitle
\begin{center}
    \vspace{-5mm}
    \centering
    \captionsetup{type=figure}
    \includegraphics[width=0.95\textwidth]{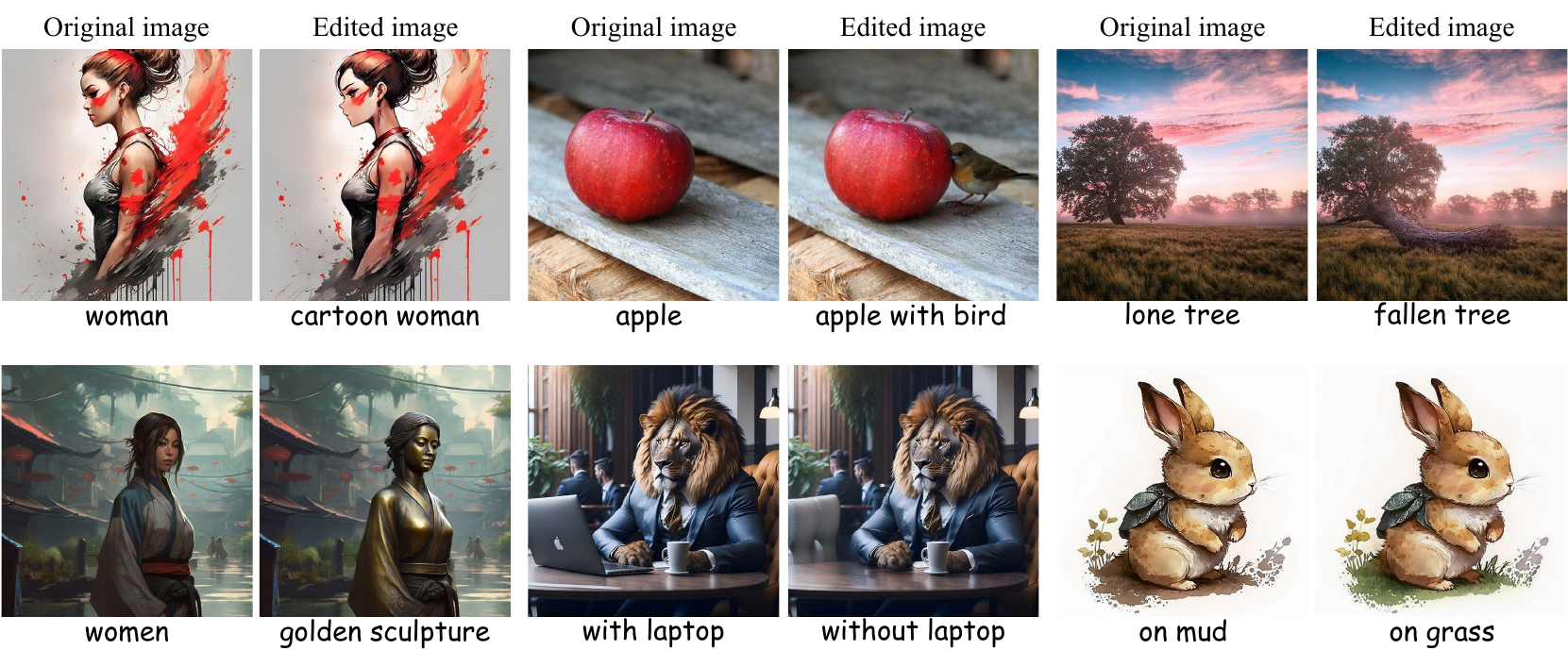}
    \caption{\textbf{Qualitative results of the proposed Conditional Velocity Correction (CVC)}. Our inversion-free, optimization-free and model agnostic CVC enables precise, fine-grained image editing across diverse tasks, including  style transfer, element addition/removal, attribute modification, and material/state conversion. CVC maintains high structural fidelity to the original image while accurately conforming to the target textual prompts.}
    \label{fig:teaser}
\end{center}%
}]
\maketitle
\vspace{-2mm}
\begin{abstract}
Recent methods in flow-based diffusion editing have enabled direct transformations between source and target image distribution without explicit inversion. However, the latent trajectories in these methods often exhibit accumulated velocity errors, leading to semantic inconsistency and loss of structural fidelity.
We propose \textbf{Conditioned Velocity Correction (CVC)}, a principled framework that reformulates flow-based editing as a distribution transformation problem driven by a known source prior.
CVC rethinks the role of velocity in inter-distribution transformation by introducing a dual-perspective velocity conversion mechanism. This mechanism explicitly decomposes the latent evolution into two components: a structure-preserving branch that remains consistent with the source trajectory, and a semantically-guided branch that drives a controlled deviation toward the target distribution.
The conditional velocity field exhibits an absolute velocity error relative to the true underlying distribution trajectory, which inherently introduces potential instability and trajectory drift in the latent space.
To address this quantifiable deviation and maintain fidelity to the true flow, we apply a posterior-consistent update to the resulting conditional velocity field. This update is derived from Empirical Bayes Inference and Tweedie correction, which ensures a mathematically grounded error compensation over time.
Our method yields stable and interpretable latent dynamics, achieving faithful reconstruction alongside smooth local semantic conversion. Comprehensive experiments demonstrate that CVC consistently achieves superior fidelity, better semantic alignment, and more reliable editing behavior across diverse tasks.
\end{abstract}
\vspace{-3mm}

\vspace{-3mm}
\section{Introduction}
\vspace{-1mm}
Diffusion models have achieved remarkable success across a wide range of generative tasks, including image~\cite{nichol2021glide,LDM,esser2024scaling,wu2025qwen}, video~\cite{kong2024hunyuanvideo,ma2025step}, 3D~\cite{poole2022dreamfusion,liu2023zero}, and even music~\cite{huang2023noise2music} generation.
In image synthesis, the most common paradigm is text-to-image generation, where a textual prompt guides the model to create visually consistent content.
Building upon this foundation, text-guided image editing~\cite{hertz2022prompt,directinv,li2025dci,kulikov2025flowedit,xie2025dnaedit} has emerged as an equally important research direction, enabling users to modify existing images according to new textual instructions.

Most image editing methods with diffusion models follow a similar process.
First, the source image is inverted into its corresponding noise representation, which is usually called inversion.
Then, this noise is gradually changed during the generation process to achieve the desired modification.
Some approaches perform editing by directly manipulating the model’s attention maps (e.g., Prompt-to-Prompt~\cite{hertz2022prompt}, MasaCtrl~\cite{masactrl}), while the others employ spatial masks~\cite{couairon2022diffedit} or adapters~\cite{ye2023ip} to localize edits and preserve identity. These approaches have proven effective within the U-Net-based diffusion architecture.

Recently, Diffusion Transformers (DiTs)~\cite{peebles2023scalable} have become the new backbone for generative models, showing impressive scalability and strong cross-modal capability.
Early DiT-based editing methods still follow the same inversion-editing process.
For instance, Stable-Flow~\cite{avrahami2025stable} first inverts the source image into noise via inversion process. Then, it injects the attention features of the guidance into the key layers of the DiT model. RF-Inversion~\cite{rout2024semantic} achieves image inversion and editing by constructing controlled forward Ordinary Differential Equations (ODEs) and reverse ODEs.
However, the inversion step in such pipelines face fundamental challenge: mapping an image back to its noise representation and reconstructing it often leads to degraded fidelity~\cite{NTI}. Although numerous subsequent works~\cite{wallace2023edict,li2025dci,rout2024semantic} have been proposed to address this limitation, they theoretically cannot attain the lossless inversion.

To overcome this limitation, FlowEdit proposes a new paradigm that avoids explicit inversion and instead constructs a direct transport path between the source and target distributions. 
Rather than repeatedly mapping between image and noise, FlowEdit defines a continuous flow path that moves directly from the source to the target manifold.
Nevertheless, this direct transformation introduces new challenges: \textbf{the intermediate latent states between the source and the target are not matched. This issue originates from an inherent limitation of FlowEdit, which applies different guidance strengths to the source and target branches. The mismatch in guidance scales introduces an unavoidable and non-negligible error from the unconditional term.
Moreover, it leads to accumulated velocity errors over time and ultimately causes structural distortion or unwanted edits in the final result.}

\vspace{-1mm}
To address these issues, we introduce \textbf{Conditioned Velocity Correction (CVC)}, a principled framework that redefines flow-based image editing through physically interpretable and mathematically consistent velocity modeling.
Our key insight is to solve the structural distortions observed in prior flow-editing methods originate from accumulated velocity errors during transport between the source and target distributions.
Instand of merely blending source and target velocities, CVC reconstructs the latent velocity field from two complementary perspectives: one branch preserves the structural consistency of the source image, while the other captures semantic deviation toward the target prompt.
This dual-branch design ensures that the editing process remains faithful to the ideal manifold while achieving controlled semantic transformation.
To further address the inherent bias in velocity field estimation, we reformulate velocity computation as an inverse problem under Empirical Bayes Inference, conditioned on the known source prior.
Within this formulation, residual trajectory errors are adaptively corrected through a Tweedie-based posterior refinement.
We regularize the integration process with an alignment loss that penalizes discrepancies between the predicted and corrected flows, enabling iterative refinement of the semantic path and ensuring both geometric fidelity and semantic precision.
Through this formulation, CVC bridges the gap between the empirical stability of flow-based editing and the theoretical consistency of diffusion dynamics.
It achieves both faithful reconstruction and smooth semantic transitions, allowing image edits to remain visually coherent, semantically aligned, and physically grounded within the diffusion manifold.

Our main contributions are summarized as follows:
\begin{itemize}
\item We provide a new interpretation of velocity misalignment in flow-based image editing, revealing that accumulated unconditional errors are the key cause of structural distortion.
\item We propose Conditioned Velocity Correction (CVC), which consists of two components: a dual-perspective velocity computation module that jointly models structural preservation and semantic deviation, and a posterior-consistent velocity correction module that refines the reconstructed flow using Tweedie-based Empirical Bayes Inference.
\item Comprehensive experiments demonstrate that CVC achieves higher reconstruction accuracy and more stable semantic transitions compared to existing flow-editing baselines, validating its theoretical soundness and practical effectiveness.
\end{itemize}
\vspace{-2mm}

\vspace{-1mm}
\section{Related Work}
\subsection{Image Generation and Editing}
Image generation has made rapid progress in recent years. Early GAN-based methods, such as GAN~\cite{goodfellow2014generative} and StyleGAN~\cite{karras2019style,karras2020analyzing}, present the foundation for controllable and high-quality image synthesis. The development of diffusion models~\cite{ho2020denoising,DDIM} further push the field forward, improving resolution, realism, and diversity. Representative models like GLIDE~\cite{nichol2021glide}, Imagen~\cite{saharia2022photorealistic}, DALL·E 2~\cite{ramesh2022hierarchical}, and Stable Diffusion~\cite{LDM} achieve remarkable performance on many generation tasks. Based on these models, a lot of diffusion-based editing methods have been proposed. Prompt-to-Prompt (P2P)~\cite{hertz2022prompt} controls image edits by adjusting cross-attention maps. Pix2Pix-Zero ~\cite{parmar2023zero} performs zero-shot image translation using text guidance. Plug-and-Play~\cite{plugandplay}, MasaCtrl~\cite{masactrl}, and IP-Adapter~\cite{ye2023ip} improve spatial or visual control through modular attention mechanisms.

Recently, Diffusion Transformer (DiT) models such as Stable Diffusion 3 (SD3)~\cite{esser2024scaling}, FLUX~\cite{flux}, FLUX-Kontext~\cite{batifol2025flux}, Bagel~\cite{zhang2025unified} and Qwen-Image~\cite{wu2025qwen} and have become the mainstream architecture, offering better scalability and semantic consistency. At the same time, flow-based inversion and editing methods~\cite{avrahami2025stable,rout2024semantic,kulikov2025flowedit,xie2025dnaedit} provide new ways to edit images.
\subsection{Inverse Problem with Diffusion Models}
Inverse Problems are fundamental tasks in science and engineering. Essentially, they aim to infer the causes or unknown parameters from observed results or effects. Inverse problems have widespread and critical applications in fields such as medical imaging (\emph{e.g.,} X-ray Computed Tomography (CT)~\cite{song2021solving}, Magnetic Resonance Imaging (MRI)~\cite{song2021solving}), geophysics~\cite{virieux2009overview}, and signal processing~\cite{boche2025inverse}.

Solving inverse problems with diffusion models has become increasingly popular. 
Thanks to their great ability in image generation, these models are now employed as powerful structural priors in various image processing tasks, such as image inpainting~\cite{chung2022improving,chung2022diffusion}, image restoration~\cite{zhu2023denoising}, image fusion~\cite{zhao2023ddfm}, super-resolution~\cite{song2023pseudoinverse}, deblurring~\cite{chung2022diffusion}. Diffusion models have not been used to solve inverse problems in image editing, and our work aims to bridge this gap.

\vspace{-3mm}
\section{Preliminary}
\noindent\textbf{Diffusion Models and Flow Matching.} Diffusion Models define a fixed forward process and a learned reverse process. Denoising Diffusion Probabilistic Models (DDPM)~\cite{ho2020denoising} rely on the learned noise predictor $\boldsymbol{\epsilon}_\theta(\mathbf{x}_t, t)$ to perform iterative sampling from the prior distribution $\mathbf{x}_T \sim \mathcal{N}(\mathbf{0}, \mathbf{I})$ down to the data sample $\mathbf{x}_0$. The sampling is typically a stochastic Markov chain defined by:$$\mathbf{x}_{t-1} {=} \frac{1}{\sqrt{\alpha_t}} \left( \mathbf{x}_t - \frac{1 - \alpha_t}{\sqrt{1 - \bar{\alpha}_t}} \boldsymbol{\epsilon}_\theta(\mathbf{x}_t, t) \right) + \sigma_t \mathbf{z},\ \mathbf{z} \sim \mathcal{N}(\mathbf{0}, \mathbf{I})$$where $\alpha_t$ and $\bar{\alpha}_t$ are variance schedules, and $\sigma_t$ controls the added noise.
On the other hand, Flow Matching (FM)~\cite{lipman2022flow}, particularly the Rectified Flow~\cite{liu2022flow} approach, avoids the SDE formulation and directly seeks a deterministic Ordinary Differential Equation (ODE) that defines the transport from the prior $\mathbf{x}_0 \sim \mathcal{N}(\mathbf{0}, \mathbf{I})$ to the data $\mathbf{x}_1 \sim p_{\text{data}}$:$$\frac{d\mathbf{x}_t}{dt} = \mathbf{v}_\theta(\mathbf{x}_t, t)$$The core idea is to learn a vector field $\mathbf{v}_\theta(\mathbf{x}, t)$ that minimizes the objective based on simple, predefined probability paths. Unlike the stochastic sampling of the reverse SDE, generation in FM is achieved by deterministic numerical integration of the learned ODE: $\mathbf{x}_{t + \Delta t} \approx \mathbf{x}_t + \Delta t \cdot \mathbf{v}_\theta(\mathbf{x}_t, t).$ This deterministic and often near-straight-line path enables the use of simple solvers like the Euler method with large step sizes, resulting in significantly faster and more efficient sampling compared to standard DDPMs.

\noindent\textbf{Image Editing.} Diffusion-based image editing typically follows the \emph{inversion-editing} paradigm, which projects an image into the noise space and then performs editing during the reverse denoising process.  
This can be interpreted as transporting samples from the source distribution to the target distribution through the Gaussian manifold. FlowEdit~\cite{kulikov2025flowedit} reformulates this process as a direct flow between the source and target domains.  
Given the forward trajectories $Z^{\text{src}}_t$ and $Z^{\text{tar}}_t$, they define a coupling path:
\begin{equation}
Z^{\text{inv}}_t = Z^{\text{src}}_0 + Z^{\text{tar}}_t - Z^{\text{src}}_t,
\end{equation}
which starts at $Z^{\text{inv}}_1 = X^{\text{src}}$ and ends at $Z^{\text{inv}}_0 = X^{\text{tar}}$.  
Differentiating w.r.t.~time yields an ODE:
\begin{equation}
dZ^{\text{inv}}_t = V^{\Delta}_t(Z^{\text{src}}_t, Z^{\text{tar}}_t)\,dt, \quad 
V^{\Delta}_t = V_2 - V_1,
\end{equation}
where $V_1$ and $V_2$ denote the velocity fields of the source and target conditions.

\vspace{-2mm}
\section{Method}
\begin{figure*}
    \centering
    \includegraphics[width=1\linewidth]{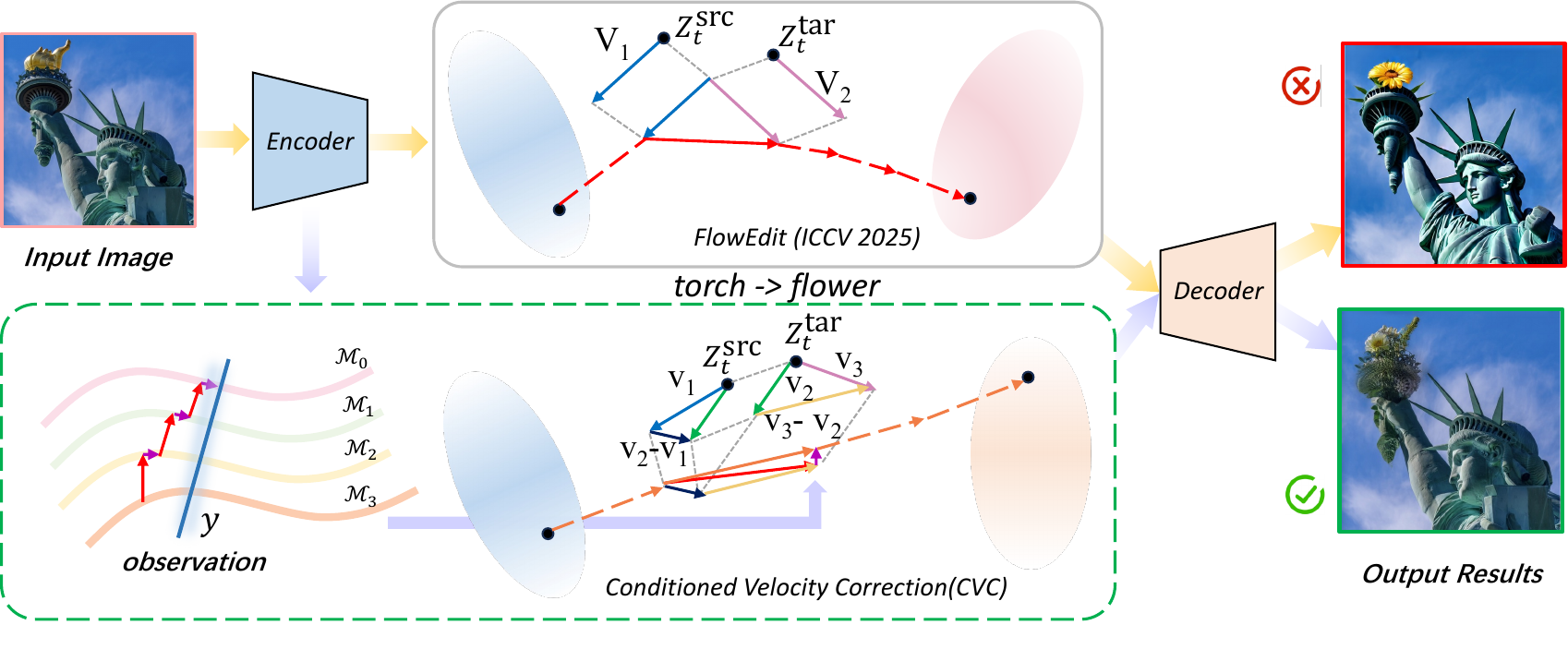}
    \caption{\textbf{FlowEdit vs. Conditioned Velocity Correction(CVC).} The upper part illustrates FlowEdit. At each time step $t$, it derives the final latent transfer velocity by simply taking the difference between the velocities corresponding to the source and target distributions. This direct subtraction inevitably leads to cumulative velocity errors over time. The bottom part presents our proposed Conditioned Velocity Correction (CVC) framework. The core innovation of CVC is that it not only significantly reduces the computational velocity error at each step but also integrates an Absolute Error Correction mechanism across the trajectory. As depicted by the orange path, our method achieves an error-free distribution transfer, ensuring an exact and consistent editing result from the source to the target distribution.}
    \label{fig:method}
    \vspace{-2mm}
\end{figure*}

\noindent\textbf{Path Integration Error.} Although FlowEdit provides an intuitive way to perform image editing by transporting samples from the source distribution to the target distribution, its results often fall short of expectations.
To better understand the underlying cause, we rethink their formulation. The velocity field in their method is computed according to the following equation:
\begin{equation}
\begin{aligned}
    V_1 &= \mathbf{\epsilon}_{\theta}(\mathbf{z}_t^{\text{src}}, t, \varnothing) + \omega_1 \cdot \left( \mathbf{\epsilon}_{\theta}(\mathbf{z}_t^{\text{src}}, t, c_{\text{}}^{\text{src}}) - \mathbf{\epsilon}_{\theta}(\mathbf{z}_t^{\text{src}}, t, \varnothing) \right) \\
    V_2 &= \mathbf{\epsilon}_{\theta}(\mathbf{z}_t^{\text{tar}}, t, \varnothing) + \omega_2 \cdot \left( \mathbf{\epsilon}_{\theta}(\mathbf{z}_t^{\text{tar}}, t, c_{\text{}}^{\text{tar}}) - \mathbf{\epsilon}_{\theta}(\mathbf{z}_t^{\text{tar}}, t, \varnothing) \right)
\end{aligned}
\end{equation}

The velocity difference is calculated as:
\begin{equation}
\begin{aligned}
V_t^{\Delta} &= V_2 - V_1 \\
&=
\left[
\underbrace{
\omega_2 \epsilon_{\theta}(\mathbf{z}_t^{\mathrm{tar}}, t, c^{\mathrm{tar}})
- \omega_1 \epsilon_{\theta}(\mathbf{z}_t^{\mathrm{src}}, t, c^{\mathrm{src}})
}_{\text{conditional term}}
\right]  \\
&\quad+
\left[
\underbrace{
(1-\omega_2)\epsilon_{\theta}(\mathbf{z}_t^{\mathrm{tar}}, t, \varnothing)
-(1-\omega_1)\epsilon_{\theta}(\mathbf{z}_t^{\mathrm{src}}, t, \varnothing)
}_{\text{unconditional term}}
\right]
\end{aligned}
\label{eq:delta_v_expanded}
\end{equation}
Generally, $z_t^{\text{src}}$ is not equal to $z_t^{\text{tar}}$, the noise prediction corresponding to the unconditional branch does not necessarily represent the desired direction of the velocity field.
Given $V^{\Delta}_{t} = V_2 - V_1$, the integration of $V$ over time can be roughly divided into two components:
the first part corresponds to the intended path deviation between the source and target trajectories, while the second part originates from the error introduced by the unconditional term.
Although this unconditional error is relatively small at each individual step, it accumulates over time and can eventually lead to noticeable distortions in the final edited result.
\begin{equation}
\begin{aligned}
x^{\Delta}_{t}
&= \int_{t_0}^{t_1} V^{\Delta}_{t}\, dt \\
&= \int_{t_0}^{t_1}
\Big[
\omega_2 \mathbf{\epsilon}_{\theta}(\mathbf{z}_t^{\text{tar}}, t, c^{\text{tar}})
- \omega_1 \mathbf{\epsilon}_{\theta}(\mathbf{z}_t^{\text{src}}, t, c^{\text{src}})
\Big] dt +\\
&\quad  \int_{t_0}^{t_1}
\Big[
(1-\omega_2)\mathbf{\epsilon}_{\theta}(\mathbf{z}_t^{\text{tar}}, t, \varnothing)
- (1-\omega_1)\mathbf{\epsilon}_{\theta}(\mathbf{z}_t^{\text{src}}, t, \varnothing)
\Big] dt \, \\
&= {x_1}^{\Delta}_{t}+{x_2}^{\Delta}_{t}
\end{aligned}
\label{eq:int_delta_v_continuous}
\end{equation}
Where ${x_1}^{\Delta}_{t}$ represents the path deviation between the source and target trajectories, ${x_2}^{\Delta}_{t}$ represents the error introduced by the unconditional term.

Based on this observation, our goal is to design a method that is both physically interpretable and mathematically consistent for updating the latent representations in flow-based image editing.
The key idea is to correct the trajectory errors of latent codes when transitioning between the source and target distributions.
This perspective allows the editing process to be formulated as a continuous semantic flow between two initial conditions.

\noindent\textbf{Conditioned Velocity Prior.} We first note that each denoising step in diffusion models uses Classifier-Free Guidance (CFG)~\cite{ho2022classifier} to better align the generated content with the intended semantics. CFG is a widely used technique in diffusion-based generative models that combines unconditional and conditional predictions to control how strongly the output follows a given condition. 
Under CFG, the conditional branch injects prompt-dependent information, while the unconditional branch serves as a neutral reference that preserves the model’s prior over natural images.

In practice, the semantic modification in an editing task is primarily driven by the \emph{conditional term}, which determines the direction and magnitude of the velocity toward the target distribution.
Eq.~\ref{eq:int_delta_v_continuous} analyzes the error caused by \emph{unconditional terms}.
A seemingly simple strategy to solve the trajectory error mentioned before is to discard the unconditional prediction and use only the conditional velocity.
Yet doing so breaks the balance. Without the unconditional component, the model becomes overly sensitive to the target condition, causing the latent code to deviate from the structural prior encoded in the source image.
As a result, edits become semantically stronger but drift away from the original geometry, leading to loss of fine-grained details, texture inconsistencies, or even global structural collapse.
To address this, we design a new way to compute the velocity difference, formulated as follows:
\begin{align}
V &\mathrel{\mathop:}= \epsilon_\theta(z, t , c),\\
v_1 &= V(z^{\mathrm{src}}_{t_i}, t_i,C_\mathrm{src}),  \\
v_2 &= V(z^{\mathrm{tar}}_{t_i}, t_i,C_\mathrm{src}),  \\
v_3 &= V(z^{\mathrm{tar}}_{t_i}, t_i,C_\mathrm{tar}). \\
V^{\Delta}_{t_i}
&= 
\alpha\,\big(v_2 - v_1\big)
+ 
\beta\,\big(v_3 - v_2\big),
\label{eq:fourquery}
\end{align}

Here, $\alpha$ and $\beta$ are two hyperparameters.
Specifically, we sample velocity information from two complementary perspectives. The first focuses on maintaining structural consistency with the source image, while the second captures semantic variation along the target path. This design preserves the structure of the original image while achieving controllable semantic transformation. 
This design enables the editing process to remain faithful to the source image, preserving its visual structure while achieving controllable semantic transformation.

With our formulation, the unconditional term is directly removed in theory. Therefore the error introduced by the unconditional term in Eq.~(\ref{eq:int_delta_v_continuous}) is eliminated, effectively resolving the first issue.
However, in practice, a residual velocity discrepancy remains due to imperfect alignment between the source ($z_{\text{src}}$) and target ($z_{\text{tar}}$) latent codes.
When $C_{\text{src}} = C_{\text{tar}}$, the formulation reaches its boundary condition, where the editing task degenerates into pure reconstruction.

However, a residual velocity difference persists, reflecting the imperfect alignment between the latent trajectories.
This error term can be simply described as follows:
\begin{equation}
V^{\Delta}_{t_i}
= \alpha\,\big(v_2 - v_1\big)= V(z^{\mathrm{tar}}_{t_i}, t_i,C_\mathrm{src}) - V(z^{\mathrm{src}}_{t_i}, t_i,C_\mathrm{src})
\label{eq:velocity_combination}
\end{equation}

To mitigate this, we exploit the known source prior $x^{0}_{\text{src}}$ as a reference to correct the residual error. With this prior, the velocity update can be conditioned. 
With this reference, the error can be reformulated as:
\begin{gather}
V^{\Delta}_{t_i}
= \bigl[(V(z^{\mathrm{tar}}_{t_i}, t_i, C_{\mathrm{src}})
- V(z^{\mathrm{src}}_{t_i}, t_i, C_{\mathrm{src}})) \mid x^{0}_{\mathrm{src}}\bigr],\\
\Delta x = V^{\Delta}_{t_i} \, \Delta t,
\label{eq:basic_dynamics}
\end{gather}
Since $t$ is defined within the range 0 to 1, the case of $\Delta t$ = 1 corresponds to the cumulative displacement $\Delta x$ experienced by the velocity $V^{\Delta}_{t_i}$.
The discrepancy between this predicted displacement and the source latent $x^{0}_{\mathrm{src}}$ represents the absolute reconstruction error in the formulation.
Our objective is to minimize the absolute error to ensure that the edited result remains faithful to the original image.

\noindent\textbf{Inverse Problem Formulation.} We formulate this process as an inverse problem with prior knowledge, where the goal is to recover the corrected latent representation under the constraint of the known source prior.
To achieve this, we employ the empirical Bayes framework with Tweedie's formula, enabling a principled estimation of the posterior and an efficient correction of the error within the diffusion process.
Defining the target distribution $p_0$ over the latent code, one can introduce a conditional flow $\psi_t(z\!\mid\!z_0)$ with velocity field $v_t(z\!\mid\!z_0)$.
The denoised estimate of $z_t$ takes the score-corrected form.
\begin{equation}
\tilde z_{0\mid t} \;=\; \hat z_{0\mid t}\;-\;\,\nabla_{\hat z_{0\mid t}}\log p\!\left(y\,\middle|\,\hat z_{0\mid t}\right).
\label{eq:z0_t_update}
\end{equation}

Given that the known source latent $x^{0}_{\mathrm{src}}$ provides an explicit reconstruction prior,
we reformulate the velocity update by conditioning only on $x^{0}_{\mathrm{src}}$.
Specifically, the instantaneous velocity is expressed as:
\begin{equation}
V^{\Delta}_{t_i}
\;\equiv\;
-\,\,
\nabla_{z}\log p\!\left(x^{0}_{\mathrm{src}}\,\middle|\,z\right)
\Big|_{z=z_{t_i}},
\end{equation}

The latent displacement will be identified with Eq.~\eqref{eq:basic_dynamics}:
\begin{equation}
\Delta x 
= -\,\,
\nabla_{z}\log p\!\left(x^{0}_{\mathrm{src}}\,\middle|\,z\right)
\Big|_{z=z_{t_i}}\,\Delta t,
\label{eq:dx_from_x0src}
\end{equation}

\begin{algorithm}[t]
  \caption{algorithm for \textbf{Conditioned Velocity Correction(CVC)}}
  \label{alg:ours_alg}
  \begin{algorithmic}[1]
    \State \textbf{Input:} input image $x^{0}_{\text{src}}$, time $\{t_i\}_{i=0}^{T}$, $n_{\text{max}}$, guidance scales $\alpha,\beta$, $C_{\text{src}},C_{\text{tar}}$, weight $\eta$.
    \State \textbf{Output:} edited image $X^{\text{tar}}$
    \State \textbf{Init:} $Z^{\text{ours}}_{t_{\text{max}}}\gets X^{\text{src}}_{0}$
    \For{$i = n_{\text{max}}$ \textbf{to} $1$}
      \State $N_{t_i}\sim\mathcal{N}(0,1)$ 
      \State $Z^{\text{src}}_{t_i}\gets (1-t_i)\,X^{\text{src}} + t_i\,N_{t_i}$
      \State $Z^{\text{tar}}_{t_i}\gets Z^{\text{edit}}_{t_i}+Z^{\text{src}}_{t_i}-X^{\text{src}}$
      \State $v_1\gets V\!\left(Z^{\text{src}}_{t_i}, t_i, C_{\text{src}}\right)$
      \State $v_2\gets V\!\left(Z^{\text{tar}}_{t_i}, t_i, C_{\text{src}}\right)$
      \State $v_3\gets V\!\left(Z^{\text{tar}}_{t_i}, t_i, C_{\text{tar}}\right)$
      \State $V^{\Delta}_{t_i}\gets \alpha\,(v_2-v_1)+\beta\,(v_3-v_2)$
      \State $V^{\Delta}_{t_i}\gets \big[\,V^{\Delta}_{t_i}\mid x^{0}_{\text{src}}\,\big]$ \Comment{condition on known prior}
      \State $\Delta x \gets V^{\Delta}_{t_i}\,(t_{i-1}-t_i)$
      \State $\mathcal{L}_{\text{align}}\gets \|\Delta x-x^{0}_{\text{src}}\|_2^2$
      \State $V^{\text{new}}\gets \Delta V-\eta\,\nabla_{\Delta V}\mathcal{L}_{\text{align}}$
      \State $Z^{\text{edit}}_{t_{i-1}}\gets Z^{\text{edit}}_{t_i}+\eta\,(t_{i-1}-t_i)\,V^{\text{new}}$
    \EndFor
    \State \textbf{Return:} $Z^{\text{edit}}_{0}=X^{\text{tar}}_{0}$
  \end{algorithmic}
\end{algorithm}

Where the gradient represents the posterior-consistent correction direction induced by Tweedie's formula under the source prior. 
Eq.~\eqref{eq:dx_from_x0src} directly estimates the semantic flow driven by the source image prior $x^{0}_{\mathrm{src}}$.
In this formulation, $\Delta x$ quantifies the cumulative latent correction, and minimizing its deviation from $x^{0}_{\mathrm{src}}$ ensures faithful reconstruction during the editing process.

\begin{figure*}[t]
    \centering
    \includegraphics[width=1\linewidth]{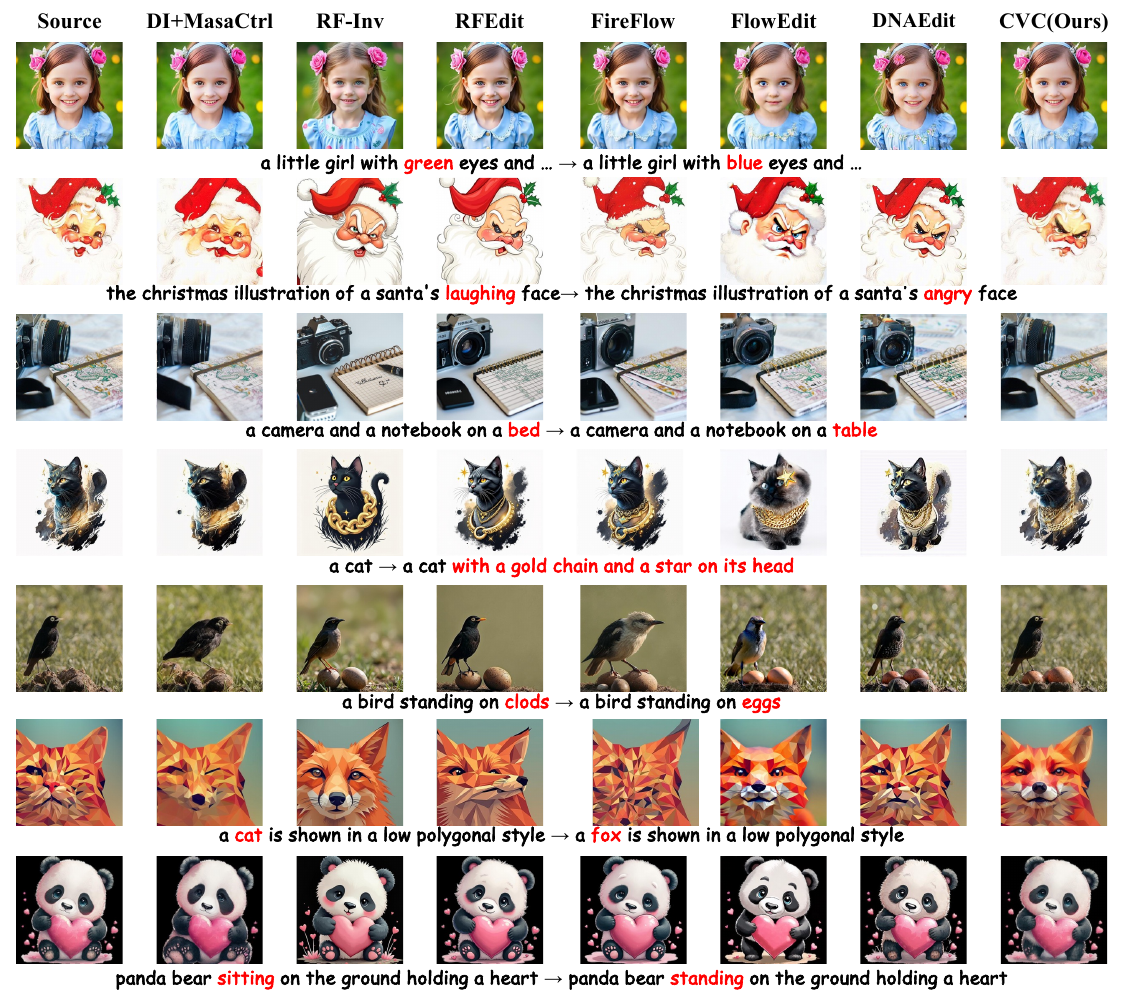}
    \vspace{-0.1in}
    \caption{\textbf{Visual results of different methods on PIE-Bench.} Each method is identified at the top of its respective column, while detailed editing information appears beneath each corresponding row. CVC(ours) demonstrates significant enhancements over existing methods.}
    \label{fig:comparison}
\end{figure*}

\paragraph{Optimization Objective.}
To ensure that the latent displacement $\Delta x$ follows the underlying semantic flow $\Delta V$, we minimize their deviation in the least-squares sense:
\begin{equation}
\mathcal{L}_{\text{align}} = 
\left\| \Delta x - x^0_\mathrm{src} \right\|_2^2.
\label{eq:align_loss}
\end{equation}

Eq.~\eqref{eq:align_loss} penalizes the discrepancy between the two latent states. A one-step gradient-based correction is then applied to refine the velocity:
\begin{equation}
{V_{new}} = \Delta V - \eta \cdot \nabla_{\Delta V} \mathcal{L}_{\text{align}} .
\end{equation}
where \( \eta \) is a hyperparameter that controls the correction strength. This update adjusts the predicted noise in a direction that reduces its divergence from the reference signal, effectively grounding the velocity in visual structure. As a result, this correction improves reconstruction fidelity and ensures that the trajectory remains semantically and perceptually consistent with the original image.
The complete process of \textbf{Conditioned Velocity Correction(CVC)} is summarized in Algorithm~\ref{alg:ours_alg}.

\section{Experiments}
\begin{table*}[t]
\caption{Performance comparison of diverse editing methods under PIE-Bench. Metrics include DINO ($\downarrow$), PSNR ($\uparrow$), LPIPS ($\downarrow$), MSE ($\downarrow$), SSIM ($\uparrow$), CLIP ($\uparrow$) and Average Ranks($\downarrow$). Best and second-best results are highlighted in \textbf{bold} and \underline{underlined}, respectively. CVC (ours) shows the best performance across most metrics, achieving the highest average ranking.}
\vspace{0.1in}

    \label{tab:metrice_edit}
    \centering
    \begin{tabular}{l|c|c|c|c|c|c|c|c}
   \toprule
   \makecell[c]{Method}  &\makecell[c]{Model}& \makecell[c]{DINO$\downarrow$ \\ $\times10^3$} & PSNR$\uparrow$ & \makecell[c]{LPIPS$\downarrow$ \\ $\times10^3$} & \makecell[c]{MSE$\downarrow$ \\ $\times10^4$} & \makecell[c]{SSIM$\uparrow$ \\ $\times10^2$} & \makecell[c]{CLIP}$\uparrow$  &\makecell[c]{Ranks}$\downarrow$ \\
   \midrule
PnP\cite{plugandplay}  &SD1.5&27.35  &22.31  &112.76  &82.95  &79.25  &25.41&10.5 \\
MasaCtrl\cite{masactrl}  &SD1.4&27.12  &22.19  &105.44  &86.37   &79.91  &24.03 &11 \\
DI+PnP\cite{directinv}  &SD1.5&23.35  &22.46  &105.51  &79.94  &79.88  &25.49&8.67 \\   
DI+MasaCtrl\cite{directinv}  &SD1.4&  23.58&22.68  &87.41  &80.63  &81.51  &24.39&8.67 \\
InfEdit\cite{xu2023inversion} &LCM& 19.31 & 27.31 &  56.32 &  47.80 & 85.30 & 24.90 &5.17\\
InsP2P\cite{brooks2023instructpix2pix}  &InsP2P &58.13 &20.80 &159.23 &221.3 &76.47 &23.61&14.5 \\
        \midrule
RF-Inv\cite{rout2024semantic}  &FLUX&42.29  & 20.20 & 179.54 & 139.2 & 69.91 & 24.57 &14\\
RFEdit\cite{wang2024taming} &FLUX&21.79 &24.83 &113.15 &52.46 &83.38 &25.57&7.17 \\
FireFlow\cite{deng2024fireflow} &FLUX&29.03 &23.33 &133.40 &70.83 &81.22 &\underline{26.19} &8.67\\
FlowEdit\cite{kulikov2025flowedit} &FLUX&27.82 &21.96 &112.19 &94.99 &83.08 &25.25&10.83\\
FlowEdit~\cite{kulikov2025flowedit}&SD3 &27.12 & 22.22 & 104.12 & 85.96 & \underline{93.22} & \textbf{27.51}&6.83 \\

DNA-edit~\cite{xie2025dnaedit}&FLUX &18.87 &24.99&95.06&50.45&85.71&25.79&4.67 \\ 
DNA-edit~\cite{xie2025dnaedit}&SD3 &14.19 &26.66&74.57&32.76&88.63&25.63 &3.83\\ \midrule
\textbf{CVC(ours)} &SD3& \underline{5.45} & \underline{31.67} & \underline{23.33} & \underline{12.27} & 93.20 & 25.64&\textbf{2.5}   \\

\textbf{CVC(ours)} &FLUX&\textbf{3.24} & \textbf{33.45} & \textbf{15.61} & \textbf{8.88} & \textbf{95.59}  & 24.50 &\underline{2.83} \\
   \bottomrule
\end{tabular}

\end{table*}
\subsection{Experimental Setups}
\label{sec:setup}
\textbf{Evaluation Metrics.} 
Our evaluation strategy for the performance of our method utilizes a comprehensive set of metrics, offering multiple aspects. Specifically, we employ the DINO score~\cite{caron2021emerging} to calculate over the entire image to capture global consistency. The CLIP score~\cite{radford2021learning} is employed to quantify the alignment between the generated image and the given prompt. Both of them are computed over the entire image. Peak Signal-to-Noise Ratio (PSNR), Mean Squared Error (MSE), Structural Similarity Index (SSIM), and Learned Perceptual Image Patch Similarity (LPIPS)~\cite{zhang2018unreasonable} are also reported. These are focused solely on evaluating local quality (background preservation and image fidelity) and are calculated strictly within the annotated regions defined by the dataset~\cite{directinv}.

\textbf{Datasets.}
The efficacy of our proposed method is primarily validated using the PIE-Bench dataset \cite{directinv}. This benchmark is structured around $700$ images, covering 10 distinct editing categories, and is rich in metadata. Specifically, each image is accompanied by five critical annotations: the source and target image prompts, an explicit editing instruction, the primary editing part and the editing mask. The computation of region-specific metrics strictly depends on the mask, ensuring evaluation is localized precisely where the editing is expected. Furthermore, we extend our performance assessment to the COCO2017 dataset \cite{lin2014microsoft} to demonstrate the robustness and applicability of our method across the real-world scenarios.

\textbf{Other Settings.} In our experiments, we utilize Stable Diffusion 3(SD3) medium~\cite{esser2024scaling} and FLUX.1 dev~\cite{flux} as the flow-based diffusion model. We set time steps as 50 in SD3 and 28 in FLUX, which are the same as those used in the baselines. For our method, we set the hyper-parameters $\alpha=1, \beta=7,\eta=0.2$ in SD3 and $\alpha=1, \beta=3.5,\eta=0.2$ in FLUX. All experiments and validations are conducted on a single NVIDIA RTX A6000 GPU.

\subsection{Comparisons with Editing Methods}
\label{sec:comparison}
\textbf{Quantitative Results.}
Table~\ref{tab:metrice_edit} presents a comprehensive comparison of our method against recent diffusion-based editing baselines under the PIE-Bench\cite{directinv}. Our method(CVC) achieves the best performance across most quantitative metrics. Specifically, it attains the lowest DINO and the highest SSIM values, indicating superior structural preservation. Also, the highest PSNR and lowest LPIPS and MSE reflect enhanced preservation. 
For fair comparison, we compare CVC-SD3 with the second-best method, DNAEdit-SD3.
CVC-SD3 achieves significant improvements, including a 61.5\% reduction in DINO(5.45vs.14.19), 18.8\% increasement in PSNR(31.67vs.26.66),  68.71\% reduction in LPIPS(23.33vs.74.57), and 62.5\% reduction in MSE(12.27vs.32.76). 
These results confirm that our framework effectively mitigates the cumulative velocity errors observed in prior flow-based formulations, producing more stable latent trajectories and visually coherent edits. Overall, CVC delivers consistent gains across both pixel-level and perceptual measures, setting a new performance benchmark for diffusion-based image editing.

\textbf{Qualitative Results.}
Figure~\ref{fig:comparison}  presents a visual comparison with other methods. The first row presents portrait images with simple attribute changes (e.g., eye color). Most methods can produce plausible transformations, yet inconsistencies appear in the face and background, whereas our CVC achieves precise editing without inconsistencies.
The second row illustrates emotion modification in illustrated characters. Competing methods often distort local geometry or fail to preserve stylistic consistency, while CVC preserves both the artistic texture and the target expression.
The third row demonstrates background editing, where previous methods struggle with objects coherence. CVC effectively transforms the surface context (“bed”→“table”) while maintaining the others unchanged.
The fourth row shows local object attribute changes; other models tend to introduce artifacts near the edited region, whereas CVC achieves faithful detail generation (e.g., gold chain, star) and preserves the overall composition.
In the fifth row, involving fine-grained contextual editing (bird and eggs), our approach maintains object boundaries and semantic alignment better than all baselines.
The sixth row evaluates cross-category style translation, where CVC produces coherent polygonal patterns without shape collapse or over-smoothing.
Finally, the last row emphasizes our method’s robustness in positional attribute transformation. CVC precisely handles pose variations (e.g., “sitting”→“standing”) while preserving structural integrity and visual quality.

\subsection{Reconstruction Comparision}
\begin{figure}[h]
    \centering
    \vspace{-4mm}
    \includegraphics[width=1\linewidth]{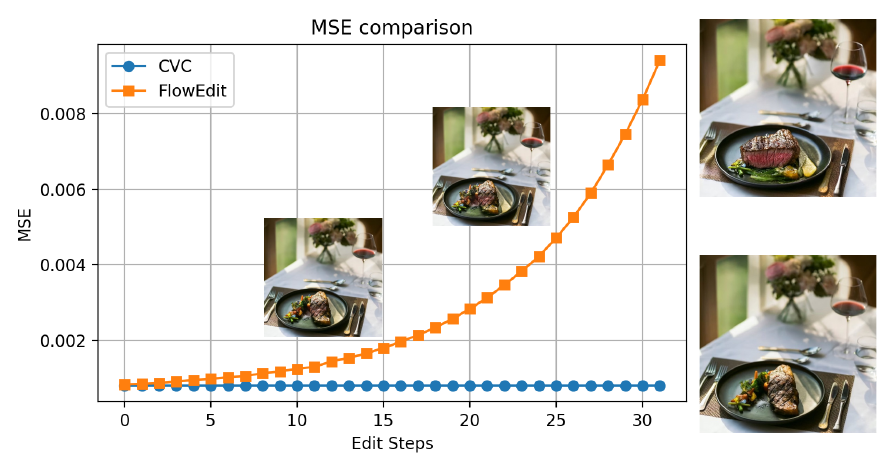}
    \vspace{-4mm}
    \caption{Reconstruction MSE via Editing Steps}
    \label{fig:rec}
\end{figure}
When the target prompt is equal to the source prompt, the image editing task degenerates into reconstruction. As discussed earlier, FlowEdit inevitably introduces an absolute reconstruction error. As shown in the Figure~\ref{fig:rec}, this error accumulates during the editing (reconstruction) process, causing the output to gradually drift away from the original image and eventually resulting in a noticeable mismatch.
In contrast, our method maintains a consistently low MSE with respect to the source image throughout the entire editing (reconstruction) trajectory. This demonstrates the stability of our formulation and highlights its effectiveness in preserving content and structural fidelity.
\subsection{Application on Style Transfer}
\begin{figure}[t]
    \centering
    \vspace{-2mm}
    \includegraphics[width=1\linewidth]{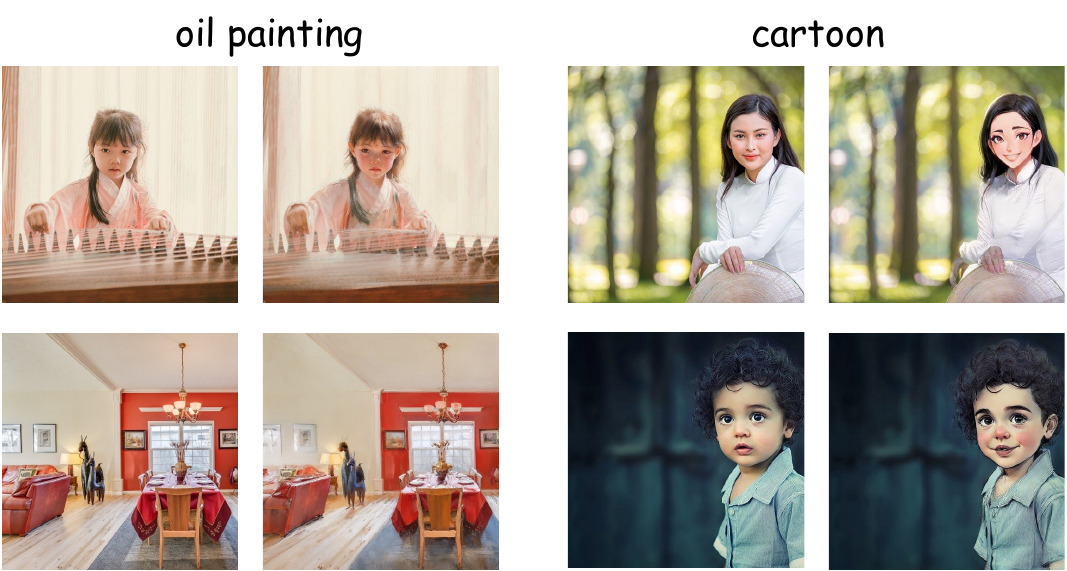}
    \caption{Application on Style Transfer}
    \label{fig:style_transfer}
    \vspace{-3mm}
\end{figure}

Our method demonstrates versatile applicability across numerous downstream tasks, with text-guided image style transfer being a prominent example. Style transfer fundamentally requires transforming an image's stylistic properties while robustly preserving its original content structure. This process is often subject to an inherent trade-off between content fidelity and style intensity. Our approach achieves superior style transfer results, which is visually substantiated in Figure~\ref{fig:style_transfer}.

\vspace{-1mm}
\subsection{Ablation Study}
\vspace{-1mm}
We conduct ablation studies on the two components of our proposed CVC framework. As shown in Table~\ref{tab:ablation}, each component provides clear performance improvement, and their combination contribute to results that are significantly better than the baseline.
\begin{table}[h]
\vspace{-2mm}
\centering
\caption{Ablation Study}
\vspace{-2mm}
\scriptsize
\setlength{\tabcolsep}{3.3pt}
\begin{tabular}{lcccccc}
\toprule
  \makecell[c]{method}&\makecell[c]{DINO$\downarrow$ \\ $\times10^3$} & PSNR$\uparrow$ & \makecell[c]{LPIPS$\downarrow$ \\ $\times10^3$} & \makecell[c]{MSE$\downarrow$ \\ $\times10^4$} & \makecell[c]{SSIM$\uparrow$ \\ $\times10^2$} & \makecell[c]{CLIP}$\uparrow$ \\
\midrule
FlowEdit &27.12 & 22.22 & 104.12 & 85.96 & 93.22 & 27.51\\
+ Dual-Perspective $v$ & 7.85 & 29.83 & 30.12 & 17.15 & 92.45 & 25.86 \\
+ $v$ +Velocity Correction &\textbf{5.45} & \textbf{31.67} & \textbf{23.33} & \textbf{12.27} & 93.20 & 25.64  \\

\bottomrule
\end{tabular}
\label{tab:ablation}
\vspace{-4mm}
\end{table}

Due to page limit, additional ablations on hyperparameters, analysis of computational cost, and experiments on more datasets are included in the supplementary material.

\vspace{-2mm}
\section{Conclusion}
\vspace{-1mm}
In this paper, we introduce \textbf{Conditioned Velocity Correction (CVC)}, an inversion-free and text-guided image editing framework built upon flow-based diffusion models.
We show that the limitation of existing approaches arises from accumulated velocity errors along the latent trajectory.
CVC addresses this by reformulating the editing process as a distribution transformation conditioned on a known source prior. By introducing a dual-perspective velocity conversion mechanism and trajectory correction, CVC achieves high structural fidelity and accurate semantic control. Extensive experiments demonstrate that CVC consistently outperforms state-of-the-art editing baselines across diverse tasks.

{
    \small
    \bibliographystyle{ieeenat_fullname}
    \bibliography{ref}

@String(NIPS= {Adv. Neural Inform. Process. Syst.})

@String(NIPS  = {NeurIPS})

@article{chung2022improving,
  title={Improving diffusion models for inverse problems using manifold constraints},
  author={Chung, Hyungjin and Sim, Byeongsu and Ryu, Dohoon and Ye, Jong Chul},
  journal={Advances in Neural Information Processing Systems},
  volume={35},
  pages={25683--25696},
  year={2022}
}

@article{chung2022diffusion,
  title={Diffusion posterior sampling for general noisy inverse problems},
  author={Chung, Hyungjin and Kim, Jeongsol and Mccann, Michael T and Klasky, Marc L and Ye, Jong Chul},
  journal={arXiv preprint arXiv:2209.14687},
  year={2022}
}

@inproceedings{zhu2023denoising,
  title={Denoising diffusion models for plug-and-play image restoration},
  author={Zhu, Yuanzhi and Zhang, Kai and Liang, Jingyun and Cao, Jiezhang and Wen, Bihan and Timofte, Radu and Van Gool, Luc},
  booktitle={Proceedings of the IEEE/CVF conference on computer vision and pattern recognition},
  pages={1219--1229},
  year={2023}
}

@inproceedings{zhao2023ddfm,
  title={DDFM: denoising diffusion model for multi-modality image fusion},
  author={Zhao, Zixiang and Bai, Haowen and Zhu, Yuanzhi and Zhang, Jiangshe and Xu, Shuang and Zhang, Yulun and Zhang, Kai and Meng, Deyu and Timofte, Radu and Van Gool, Luc},
  booktitle={Proceedings of the IEEE/CVF international conference on computer vision},
  pages={8082--8093},
  year={2023}
}

@inproceedings{song2023pseudoinverse,
  title={Pseudoinverse-guided diffusion models for inverse problems},
  author={Song, Jiaming and Vahdat, Arash and Mardani, Morteza and Kautz, Jan},
  booktitle={International Conference on Learning Representations},
  year={2023}
}

@inproceedings{esser2024scaling,
  title={Scaling rectified flow transformers for high-resolution image synthesis},
  author={Esser, Patrick and Kulal, Sumith and Blattmann, Andreas and Entezari, Rahim and M{\"u}ller, Jonas and Saini, Harry and Levi, Yam and Lorenz, Dominik and Sauer, Axel and Boesel, Frederic and others},
  booktitle={Forty-first international conference on machine learning},
  year={2024}
}

@article{batifol2025flux,
  title={FLUX. 1 Kontext: Flow Matching for In-Context Image Generation and Editing in Latent Space},
  author={Batifol, Stephen and Blattmann, Andreas and Boesel, Frederic and Consul, Saksham and Diagne, Cyril and Dockhorn, Tim and English, Jack and English, Zion and Esser, Patrick and Kulal, Sumith and others},
  journal={arXiv e-prints},
  pages={arXiv--2506},
  year={2025}
}

@article{wu2025qwen,
  title={Qwen-image technical report},
  author={Wu, Chenfei and Li, Jiahao and Zhou, Jingren and Lin, Junyang and Gao, Kaiyuan and Yan, Kun and Yin, Sheng-ming and Bai, Shuai and Xu, Xiao and Chen, Yilei and others},
  journal={arXiv preprint arXiv:2508.02324},
  year={2025}
}

@article{nichol2021glide,
  title={Glide: Towards photorealistic image generation and editing with text-guided diffusion models},
  author={Nichol, Alex and Dhariwal, Prafulla and Ramesh, Aditya and Shyam, Pranav and Mishkin, Pamela and McGrew, Bob and Sutskever, Ilya and Chen, Mark},
  journal={arXiv preprint arXiv:2112.10741},
  year={2021}
}

@article{saharia2022photorealistic,
  title={Photorealistic text-to-image diffusion models with deep language understanding},
  author={Saharia, Chitwan and Chan, William and Saxena, Saurabh and Li, Lala and Whang, Jay and Denton, Emily L and Ghasemipour, Kamyar and Gontijo Lopes, Raphael and Karagol Ayan, Burcu and Salimans, Tim and others},
  journal={Advances in Neural Information Processing Systems},
  volume={35},
  pages={36479--36494},
  year={2022}
}

@article{ramesh2022hierarchical,
  title={Hierarchical text-conditional image generation with clip latents},
  author={Ramesh, Aditya and Dhariwal, Prafulla and Nichol, Alex and Chu, Casey and Chen, Mark},
  journal={arXiv preprint arXiv:2204.06125},
  volume={1},
  number={2},
  pages={3},
  year={2022}
}

@inproceedings{LDM,
  title={High-resolution image synthesis with latent diffusion models},
  author={Rombach, Robin and Blattmann, Andreas and Lorenz, Dominik and Esser, Patrick and Ommer, Bj{\"o}rn},
  booktitle={Proceedings of the IEEE/CVF conference on computer vision and pattern recognition},
  pages={10684--10695},
  year={2022}
}

@article{hertz2022prompt,
  title={Prompt-to-prompt image editing with cross attention control},
  author={Hertz, Amir and Mokady, Ron and Tenenbaum, Jay and Aberman, Kfir and Pritch, Yael and Cohen-Or, Daniel},
  journal={arXiv preprint arXiv:2208.01626},
  year={2022}
}

@inproceedings{parmar2023zero,
  title={Zero-shot image-to-image translation},
  author={Parmar, Gaurav and Kumar Singh, Krishna and Zhang, Richard and Li, Yijun and Lu, Jingwan and Zhu, Jun-Yan},
  booktitle={ACM SIGGRAPH 2023 conference proceedings},
  pages={1--11},
  year={2023}
}

@inproceedings{plugandplay,
  title={Plug-and-play diffusion features for text-driven image-to-image translation},
  author={Tumanyan, Narek and Geyer, Michal and Bagon, Shai and Dekel, Tali},
  booktitle={Proceedings of the IEEE/CVF Conference on Computer Vision and Pattern Recognition},
  pages={1921--1930},
  year={2023}
}

@article{masactrl,
  title={MasaCtrl: Tuning-Free Mutual Self-Attention Control for Consistent Image Synthesis and Editing},
  author={Cao, Mingdeng and Wang, Xintao and Qi, Zhongang and Shan, Ying and Qie, Xiaohu and Zheng, Yinqiang},
  journal={arXiv preprint arXiv:2304.08465},
  year={2023}
}

@inproceedings{karras2019style,
  title={A style-based generator architecture for generative adversarial networks},
  author={Karras, Tero and Laine, Samuli and Aila, Timo},
  booktitle={Proceedings of the IEEE/CVF conference on computer vision and pattern recognition},
  pages={4401--4410},
  year={2019}
}

@inproceedings{karras2020analyzing,
  title={Analyzing and improving the image quality of stylegan},
  author={Karras, Tero and Laine, Samuli and Aittala, Miika and Hellsten, Janne and Lehtinen, Jaakko and Aila, Timo},
  booktitle={Proceedings of the IEEE/CVF conference on computer vision and pattern recognition},
  pages={8110--8119},
  year={2020}
}

@inproceedings{goodfellow2014generative,
  title={Generative Adversarial Nets},
  author={Goodfellow, Ian J and others},
  booktitle={NIPS},
  year={2014}
}

@article{ho2020denoising,
  title={Denoising diffusion probabilistic models},
  author={Ho, Jonathan and Jain, Ajay and Abbeel, Pieter},
  journal={Advances in neural information processing systems},
  volume={33},
  pages={6840--6851},
  year={2020}
}

@inproceedings{DDIM,
  title={Denoising Diffusion Implicit Models},
  author={Song, Jiaming and Meng, Chenlin and Ermon, Stefano},
  booktitle={International Conference on Learning Representations},
  year={2020}
}

@article{ye2023ip,
  title={Ip-adapter: Text compatible image prompt adapter for text-to-image diffusion models},
  author={Ye, Hu and Zhang, Jun and Liu, Sibo and Han, Xiao and Yang, Wei},
  journal={arXiv preprint arXiv:2308.06721},
  year={2023}
}

@article{zhang2025unified,
  title={Unified multimodal understanding and generation models: Advances, challenges, and opportunities},
  author={Zhang, Xinjie and Guo, Jintao and Zhao, Shanshan and Fu, Minghao and Duan, Lunhao and Hu, Jiakui and Chng, Yong Xien and Wang, Guo-Hua and Chen, Qing-Guo and Xu, Zhao and others},
  journal={arXiv preprint arXiv:2505.02567},
  year={2025}
}

@inproceedings{caron2021emerging,
  title={Emerging properties in self-supervised vision transformers},
  author={Caron, Mathilde and Touvron, Hugo and Misra, Ishan and J{\'e}gou, Herv{\'e} and Mairal, Julien and Bojanowski, Piotr and Joulin, Armand},
  booktitle={Proceedings of the IEEE/CVF international conference on computer vision},
  pages={9650--9660},
  year={2021}
}

@inproceedings{radford2021learning,
  title={Learning transferable visual models from natural language supervision},
  author={Radford, Alec and Kim, Jong Wook and Hallacy, Chris and Ramesh, Aditya and Goh, Gabriel and Agarwal, Sandhini and Sastry, Girish and Askell, Amanda and Mishkin, Pamela and Clark, Jack and others},
  booktitle={International conference on machine learning},
  pages={8748--8763},
  year={2021},
  organization={PMLR}
}

@inproceedings{zhang2018unreasonable,
  title={The unreasonable effectiveness of deep features as a perceptual metric},
  author={Zhang, Richard and Isola, Phillip and Efros, Alexei A and Shechtman, Eli and Wang, Oliver},
  booktitle={Proceedings of the IEEE conference on computer vision and pattern recognition},
  pages={586--595},
  year={2018}
}

@article{directinv,
  title={Direct inversion: Boosting diffusion-based editing with 3 lines of code},
  author={Ju, Xuan and Zeng, Ailing and Bian, Yuxuan and Liu, Shaoteng and Xu, Qiang},
  journal={arXiv preprint arXiv:2310.01506},
  year={2023}
}

@inproceedings{lin2014microsoft,
  title={Microsoft coco: Common objects in context},
  author={Lin, Tsung-Yi and Maire, Michael and Belongie, Serge and Hays, James and Perona, Pietro and Ramanan, Deva and Doll{\'a}r, Piotr and Zitnick, C Lawrence},
  booktitle={Computer vision--ECCV 2014: 13th European conference, zurich, Switzerland, September 6-12, 2014, proceedings, part v 13},
  pages={740--755},
  year={2014},
  organization={Springer}
}

@misc{flux,
  author = {Black Forest Labs},
  title = {{Flux}},
  year = {2024},
  howpublished = {\url{https://github.com/black-forest-labs/flux}},
  note = {Accessed: 2024-11-14}
}

@article{kong2024hunyuanvideo,
  title={Hunyuanvideo: A systematic framework for large video generative models},
  author={Kong, Weijie and Tian, Qi and Zhang, Zijian and Min, Rox and Dai, Zuozhuo and Zhou, Jin and Xiong, Jiangfeng and Li, Xin and Wu, Bo and Zhang, Jianwei and others},
  journal={arXiv preprint arXiv:2412.03603},
  year={2024}
}

@article{ma2025step,
  title={Step-video-t2v technical report: The practice, challenges, and future of video foundation model},
  author={Ma, Guoqing and Huang, Haoyang and Yan, Kun and Chen, Liangyu and Duan, Nan and Yin, Shengming and Wan, Changyi and Ming, Ranchen and Song, Xiaoniu and Chen, Xing and others},
  journal={arXiv preprint arXiv:2502.10248},
  year={2025}
}

@article{huang2023noise2music,
  title={Noise2music: Text-conditioned music generation with diffusion models},
  author={Huang, Qingqing and Park, Daniel S and Wang, Tao and Denk, Timo I and Ly, Andy and Chen, Nanxin and Zhang, Zhengdong and Zhang, Zhishuai and Yu, Jiahui and Frank, Christian and others},
  journal={arXiv preprint arXiv:2302.03917},
  year={2023}
}

@article{poole2022dreamfusion,
  title={Dreamfusion: Text-to-3d using 2d diffusion},
  author={Poole, Ben and Jain, Ajay and Barron, Jonathan T and Mildenhall, Ben},
  journal={arXiv preprint arXiv:2209.14988},
  year={2022}
}

@inproceedings{liu2023zero,
  title={Zero-1-to-3: Zero-shot one image to 3d object},
  author={Liu, Ruoshi and Wu, Rundi and Van Hoorick, Basile and Tokmakov, Pavel and Zakharov, Sergey and Vondrick, Carl},
  booktitle={Proceedings of the IEEE/CVF international conference on computer vision},
  pages={9298--9309},
  year={2023}
}

@article{li2025dci,
  title={DCI: Dual-Conditional Inversion for Boosting Diffusion-Based Image Editing},
  author={Li, Zixiang and Wang, Haoyu and Wang, Wei and Tan, Chuangchuang and Wei, Yunchao and Zhao, Yao},
  journal={arXiv preprint arXiv:2506.02560},
  year={2025}
}

@inproceedings{kulikov2025flowedit,
  title={Flowedit: Inversion-free text-based editing using pre-trained flow models},
  author={Kulikov, Vladimir and Kleiner, Matan and Huberman-Spiegelglas, Inbar and Michaeli, Tomer},
  booktitle={Proceedings of the IEEE/CVF International Conference on Computer Vision},
  pages={19721--19730},
  year={2025}
}

@article{xie2025dnaedit,
  title={DNAEdit: Direct Noise Alignment for Text-Guided Rectified Flow Editing},
  author={Xie, Chenxi and Li, Minghan and Li, Shuai and Wu, Yuhui and Yi, Qiaosi and Zhang, Lei},
  journal={arXiv preprint arXiv:2506.01430},
  year={2025}
}

@article{couairon2022diffedit,
  title={Diffedit: Diffusion-based semantic image editing with mask guidance},
  author={Couairon, Guillaume and Verbeek, Jakob and Schwenk, Holger and Cord, Matthieu},
  journal={arXiv preprint arXiv:2210.11427},
  year={2022}
}

@inproceedings{peebles2023scalable,
  title={Scalable diffusion models with transformers},
  author={Peebles, William and Xie, Saining},
  booktitle={Proceedings of the IEEE/CVF international conference on computer vision},
  pages={4195--4205},
  year={2023}
}

@inproceedings{avrahami2025stable,
  title={Stable flow: Vital layers for training-free image editing},
  author={Avrahami, Omri and Patashnik, Or and Fried, Ohad and Nemchinov, Egor and Aberman, Kfir and Lischinski, Dani and Cohen-Or, Daniel},
  booktitle={Proceedings of the Computer Vision and Pattern Recognition Conference},
  pages={7877--7888},
  year={2025}
}

@article{rout2024semantic,
  title={Semantic image inversion and editing using rectified stochastic differential equations},
  author={Rout, Litu and Chen, Yujia and Ruiz, Nataniel and Caramanis, Constantine and Shakkottai, Sanjay and Chu, Wen-Sheng},
  journal={arXiv preprint arXiv:2410.10792},
  year={2024}
}

@inproceedings{NTI,
  title={Null-text inversion for editing real images using guided diffusion models},
  author={Mokady, Ron and Hertz, Amir and Aberman, Kfir and Pritch, Yael and Cohen-Or, Daniel},
  booktitle={Proceedings of the IEEE/CVF Conference on Computer Vision and Pattern Recognition},
  pages={6038--6047},
  year={2023}
}

@inproceedings{wallace2023edict,
  title={Edict: Exact diffusion inversion via coupled transformations},
  author={Wallace, Bram and Gokul, Akash and Naik, Nikhil},
  booktitle={Proceedings of the IEEE/CVF Conference on Computer Vision and Pattern Recognition},
  pages={22532--22541},
  year={2023}
}

@article{xu2023inversion,
  title={Inversion-free image editing with natural language},
  author={Xu, Sihan and Huang, Yidong and Pan, Jiayi and Ma, Ziqiao and Chai, Joyce},
  journal={arXiv preprint arXiv:2312.04965},
  year={2023}
}

@inproceedings{brooks2023instructpix2pix,
  title={Instructpix2pix: Learning to follow image editing instructions},
  author={Brooks, Tim and Holynski, Aleksander and Efros, Alexei A},
  booktitle={Proceedings of the IEEE/CVF conference on computer vision and pattern recognition},
  pages={18392--18402},
  year={2023}
}

@article{wang2024taming,
  title={Taming rectified flow for inversion and editing},
  author={Wang, Jiangshan and Pu, Junfu and Qi, Zhongang and Guo, Jiayi and Ma, Yue and Huang, Nisha and Chen, Yuxin and Li, Xiu and Shan, Ying},
  journal={arXiv preprint arXiv:2411.04746},
  year={2024}
}

@article{deng2024fireflow,
  title={Fireflow: Fast inversion of rectified flow for image semantic editing},
  author={Deng, Yingying and He, Xiangyu and Mei, Changwang and Wang, Peisong and Tang, Fan},
  journal={arXiv preprint arXiv:2412.07517},
  year={2024}
}

@article{ho2022classifier,
  title={Classifier-free diffusion guidance},
  author={Ho, Jonathan and Salimans, Tim},
  journal={arXiv preprint arXiv:2207.12598},
  year={2022}
}

@article{lipman2022flow,
  title={Flow matching for generative modeling},
  author={Lipman, Yaron and Chen, Ricky TQ and Ben-Hamu, Heli and Nickel, Maximilian and Le, Matt},
  journal={arXiv preprint arXiv:2210.02747},
  year={2022}
}

@article{liu2022flow,
  title={Flow straight and fast: Learning to generate and transfer data with rectified flow},
  author={Liu, Xingchao and Gong, Chengyue and Liu, Qiang},
  journal={arXiv preprint arXiv:2209.03003},
  year={2022}
}

@article{virieux2009overview,
  title={An overview of full-waveform inversion in exploration geophysics},
  author={Virieux, Jean and Operto, St{\'e}phane},
  journal={Geophysics},
  volume={74},
  number={6},
  pages={WCC1--WCC26},
  year={2009},
  publisher={Society of Exploration Geophysicists}
}

@article{song2021solving,
  title={Solving inverse problems in medical imaging with score-based generative models},
  author={Song, Yang and Shen, Liyue and Xing, Lei and Ermon, Stefano},
  journal={arXiv preprint arXiv:2111.08005},
  year={2021}
}

@article{boche2025inverse,
  title={Inverse problems are solvable on real number signal processing hardware},
  author={Boche, Holger and Fono, Adalbert and Kutyniok, Gitta},
  journal={Applied and Computational Harmonic Analysis},
  volume={74},
  pages={101719},
  year={2025},
  publisher={Elsevier}
}
}

\end{document}